\documentclass[10pt,twocolumn,letterpaper]{article}

\usepackage{cvpr}              

\usepackage{graphicx}
\usepackage{amsmath}
\usepackage{amssymb}
\usepackage{booktabs}
\usepackage{amsmath}
\usepackage{multirow}
\usepackage{dsfont}
\usepackage{amsmath}
\usepackage{amssymb}
\usepackage{booktabs}
\usepackage{amsmath}
\usepackage{multirow}
\usepackage{amsfonts}
\usepackage{amsthm}
\usepackage{multirow}
\usepackage{dsfont}
\newtheorem{theorem}{Theorem}[section]

\theoremstyle{definition}
\newtheorem{definition}{Definition}[section]

\usepackage[pagebackref,breaklinks,colorlinks]{hyperref}

\usepackage[capitalize]{cleveref}
\crefname{section}{Sec.}{Secs.}
\Crefname{section}{Section}{Sections}
\Crefname{table}{Table}{Tables}
\crefname{table}{Tab.}{Tabs.}

\def\cvprPaperID{366} 
\def\confName{CVPR}
\def\confYear{2023}

\begin{document}


\title{Mind the Label Shift of Augmentation-based Graph OOD Generalization}

\author{Junchi Yu$^{1,2}$,
Jian Liang$^{1}$,
Ran He$^{1,2,3}$\thanks{Corresponding Author}\\
\and
$^1$MAIS\&CRIPAC, Institute of Automation, Chinese Academy of Sciences, China\\
$^2$School of Artificial Intelligence, University of Chinese Academy of Sciences, China\\
$^3$School of Information Science and Technology, ShanghaiTech University, China\\
{\tt\small yujunchi2019@ia.ac.cn, liangjian92@gmail.com, rhe@nlpr.ia.ac.cn}
}
\maketitle

\begin{abstract}
Out-of-distribution (OOD) generalization is an important issue for Graph Neural Networks (GNNs).
Recent works employ different graph editions to generate augmented environments and learn an invariant GNN for generalization.
However, the label shift usually occurs in augmentation since graph structural edition inevitably alters the graph label. 
This brings inconsistent predictive relationships among augmented environments, which is harmful to generalization.
To address this issue, we propose \textbf{LiSA}, which generates label-invariant augmentations to facilitate graph OOD generalization.
Instead of resorting to graph editions, LiSA exploits \textbf{L}abel-\textbf{i}nvariant \textbf{S}ubgraphs of the training graphs to construct \textbf{A}ugmented environments.
Specifically, LiSA first designs the variational subgraph generators to extract locally predictive patterns and construct multiple label-invariant subgraphs efficiently.
Then, the subgraphs produced by different generators are collected to build different augmented environments.
To promote diversity among augmented environments, LiSA further introduces a tractable energy-based regularization to enlarge pair-wise distances between the distributions of environments.
In this manner, LiSA generates diverse augmented environments with a consistent predictive relationship and facilitates learning an invariant GNN.
Extensive experiments on node-level and graph-level OOD benchmarks show that LiSA achieves impressive generalization performance with different GNN backbones. Code is available on \url{https://github.com/Samyu0304/LiSA}.

\end{abstract}


\section{Introduction}
Learning from graph-structured data is a fundamental problem in various applications, such as 3D vision \cite{tailor2021towards}, knowledge graph reasoning \cite{zhou2022leverage}, and social network analysis \cite{khanam2022homophily}. 
Recently, the Graph Neural Networks (GNNs) \cite{gcn} have become a \emph{de facto} standard in developing deep learning systems on graphs \cite{chang2022adversarial}, showing superior performance on point cloud classification \cite{fischer2021stickypillars}, recommendation system \cite{wu2020graph}, biochemistry \cite{jin2018junction} and so on. 
Despite their remarkable success, these models heavily rely on the i.i.d. assumption that the training and testing data are independently drawn from an identical distribution \cite{chang2021not,li2022out}. 
When tested on out-of-distribution (OOD) graphs (\ie larger graphs), GNN usually suffers from unsatisfactory performances and unstable prediction results. 
Hence, handling the distribution shift for GNNs has received increasing attention.

Many solutions have been proposed to explore the OOD generalization problem in Euclidean space \cite{shen2021towards}, such as invariant learning \cite{arjovsky2019invariant,krueger2021out,chen2023pareto}, group fairness \cite{kleinberg2018algorithmic}, and distribution-robust optimization \cite{sinha2017certifying}. 
Recent works mainly resort to learning an invariant classifier that performs equally well in different training environments \cite{arjovsky2019invariant,krueger2021out,creager2021environment,koyama2020out}.
However, the study of its counterpart problem for non-Euclidean graphs is comparatively lacking.
One challenge is the environmental scarcity of graph-structured data \cite{li2022out,wu2022survey}. 
Inspired by the data augmentation literature \cite{volpi2018generalizing,shorten2019survey}, some pioneering works propose to generate augmented training environments by applying different graph edition policies to the training graphs \cite{wu2022discovering,wu2022handling}.
After training in these environments, the GNN is expected to have better OOD generalization ability.
Nevertheless, the graph labels may change during the graph edition since they are sensitive to graph structural modifications.
This causes the label shift problem of augmented graphs.
For example, methods of graph adversarial attack usually seek to modify the graph structure to permute the model prediction \cite{chang2021not}.
Moreover, a small structural modification can drastically influence the biochemical property of molecule or protein graphs \cite{jin2020multi}.

We formalize the impact of the label shift in augmentations on generalization using a unified structure equation model \cite{ahuja2021invariance}.
Our analysis indicates that the label shift causes inconsistent predictive relationships among the augmented environments.
This misguides the GNN to output a perturbed prediction rather than the invariant prediction, making the learned GNN hard to generalize (see Section~\ref{section-sem} for more details).
Thus, it is crucial to generate label-invariant augmentations for graph OOD generalization. 
However, designing label-invariant graph edition is nontrivial or even brings extensive computation, since it requires learning class-conditional distribution for discrete and irregular graphs.
In this work, we propose a novel label-invariant subgraph augmentation method, dubbed \textit{LiSA}, for the graph OOD generalization problem. 
For an input graph, LiSA first designs the variational subgraph generators to identify locally predictive patterns (\ie important nodes or edges for the graph label) and generate multiple label-invariant subgraphs.
These subgraphs capture prediction-relevant information with different structures, and thus construct augmented environments with a consistent predictive relationship.
To promote diversity among the augmentations, we propose a tractable energy-based regularization to enlarge the pair-wise distances between the distributions of augmented environments.
With the augmentations produced by LiSA, a GNN classifier is learned to be invariant across these augmented environments.
The GNN predictor and variational subgraph generators are jointly optimized with a bi-level optimization scheme \cite{yu2020graph}.
LiSA is model-agnostic and is flexible in handling both graph-level and node-level distribution shifts.
Extensive experiments indicate that LiSA enjoys satisfactory performance gain over the baselines on 7 graph classification datasets and 4 node classification datasets.
Our contributions are as follows:
\begin{itemize}
    \item We propose a model-agnostic label-invariant subgraph augmentation (LiSA) framework to generate augmented environments with consistent predictive relationships for graph OOD generalization.
    \item We propose the variational subgraph generator to discover locally crucial patterns to construct the label-invariant subgraphs efficiently.
    \item To further promote diversity, we further propose an energy-based regularization to enlarge pair-wise distances between the distributions of different augmented environments.
    \item Extensive experiments on node-level and graph-level tasks indicate that LiSA enjoys satisfactory performance gain over the baselines on various backbones.
\end{itemize}

\section{Related Work}
\textbf{Graph Neural Networks.} The Graph Neural Network (GNN) has become a building block for deep graph learning \cite{gcn}. It leverages the message-passing module to aggregate the adjacent information to the central node, which shows expressive power in embedding rational data. Various GNN variants have shown superior performance on social network analysis \cite{bian2020rumor}, recommender system \cite{wu2020graph}, and  biochemistry \cite{ yu2022structure}. While GNNs have achieved notable success on many tasks, they rely on the i.i.d assumption that the training and testing samples are drawn independently from the same distribution \cite{chang2020restricted}. This triggers concerns about the applications of GNN-based models in real-world scenarios where there is a distribution shift between the training and testing data.

\textbf{Out-of-distribution (OOD) Generalization.} Given the training samples from several source domains, out-of-distribution generalization aims at generalizing deep models to unseen test environments \cite{arjovsky2019invariant}. Recent studies focus on learning an invariant predictive relationship across these training environments, such as invariant representation/predictor learning \cite{krueger2021out,arjovsky2019invariant,koyama2020out}, and invariant causal prediction \cite{buhlmann2020invariance,peters2016causal,heinze2018invariant}. 
They either learn a predictor that performs equally well (also known as equi-predictive \cite{koyama2020out}) in different environments or seek a stable parent variable of the label in the structural causal model (SCM) \cite{jiang2022invariant} for the prediction.
When the environment partition is missing, many works resort to group robust optimization \cite{qian2019robust,sagawa2019distributionally,hu2018does}, environment inference \cite{zhang2022correct,creager2021environment}, and data augmentation \cite{cubuk2018autoaugment,volpi2018generalizing}.
Although domain generalization on Euclidean data has drawn much attention, the focus on its counterpart to the graph-structured data is comparatively lacking \cite{chen2022invariance}. Our work follows the data augmentation strategy. Differently, we consider generating the augmentations of graph-structured data, which is usually challenging due to their discrete and irregular structures.

\textbf{OOD Generalization on Graphs.}  Some pioneering works \cite{baranwal2021graph,bevilacqua2021size} on graph OOD generalization study whether the GNN trained on small graphs can generalize to larger graph size \cite{chuang2022tree}. Recently, researchers have extended OOD generalization methods, such as invariant learning \cite{krueger2021out} to handle the distribution shift on graphs \cite{li2022out,chen2022invariance}. 
The main challenge is environmental scarcity, which makes the learned invariant relationship insufficient for graph OOD generalization.
To this end, recent works \cite{wu2022discovering,wu2022handling,buffelli2022sizeshiftreg} employ different graph edition policies to generate augmented environments. 
Since the graph label is sensitive to the graph structure, graph editing is prone to change the label of the augmented graph and makes it difficult for graph OOD generalization

\section{Graph Augmentation for Graph OOD Generalization}
\label{section-sem}
\subsection{Problem formulation}
Let $D_{tr}=\{(G_{i},Y_{i})|1\leq i\leq N\}$ be the training graphs which are sampled from the distribution $p(G,Y)=\sum_{e\in\mathcal{E}_{tr}}p(G,Y|e)p(e)$. 
Here, $G\in \mathcal{G}$ and $Y\in\mathcal{Y}$ are graphs and their labels. $e\in\mathcal{E}$ represents the environment. The goal of graph OOD generalization is to learn a  graph neural network (GNN) $f:\mathcal{G}\rightarrow \mathcal{Y}$ on $D_{tr}$, which can generalize to unseen testing environments. This is formulated as a bi-level optimization problem, which minimizes the worst-case risk across the training environments \cite{arjovsky2019invariant}:
\begin{equation}
\begin{aligned}
\min_{f}\mathcal{R}^{e}(f), s.t. e=\arg\max_{e\in\mathcal{E}_{tr}}\mathcal{R}^{e}(f).
\end{aligned}
\label{good}
\end{equation}
Here, $\mathcal{R}^{e}(f)$ is the risk of $f$ in environment $e$. A GNN $f$ which minimizes Eqn.~\ref{good} is called invariant GNN, and is supposed to generalize to OOD graphs at testing time. 
Recent works show that the performance drop on OOD graphs is attributed to learning from spurious subgraphs, which is unstable across different environments \cite{wu2022discovering,chen2022invariance}.
Thus, they aim to learn an invariant predictive relationship between the causal subgraph $G_{inv}$ and the graph label $Y$.
A GNN leveraging such a predictive relationship is stable across different environments and is supposed to generalize.
However, the training environments for graphs \cite{chen2022invariance,li2022out} are usually scarce, making it difficult to learn a generalizable GNN.

\subsection{Augmentation-based Graph OOD Generalization}
To address the environmental scarcity issue, some works employ different graph editing policies to change the graph structures for augmentation.
For example, EERM \cite{wu2022handling} introduces a graph extrapolation strategy, which adds new edges to the training graphs with reinforcement learning.
DIR \cite{wu2022discovering} exchanges part of graph structures within a training batch, which is known as the graph intervention strategy. 
We elaborate more details on these methods in the appendix. 
While different augmentation strategies have been proposed, they are likely to cause the label shift in augmentations since graph labels are sensitive to structure editing.
This introduces inconsistent predictive relationships among the augmented environments, and brings negative effects on graph OOD generalization.
To formalize this problem, we build a unified structural equation model (SEM) for augmentation-based graph OOD generalization:
\begin{equation}
\begin{aligned}
Y_{Aug}^{e}&\leftarrow I(W_{inv}\cdot G_{inv}^{e})\oplus I(W_{aug}\cdot G_{aug}^{e})\oplus N^{e}\\
G_{Aug}^{e}&\leftarrow S_{aug}(G^{e},G_{aug}^{e})\\
N^{e}&\sim \mathrm{Bernoulli}(q),q< 0.5\\
N^{e}&\perp(G^{e},G_{aug}^{e})
\end{aligned}
\label{scm}
\end{equation}

Here, $G_{Aug}^{e}$ and $Y_{Aug}^{e}$ are the augmented graph and its label. $S_{aug}$ is the augmentation function, which generates $G_{Aug}^{e}$ given $G^{e}$ and the augmented structure $G_{aug}^{e}$. The formulation of $S_{aug}$ depends on the augmentation strategy. For EERM, $S_{aug}$ represents appending $G_{aug}^{e}$ to $G^{e}$. And it denotes exchanging $G_{aug}^{e}$ between batched $G^{e}$ for DIR. $I(\cdot)$ is the labeling function.
$W_{inv}$ is the parameterized invariant prediction relationship within original graphs and $W_{aug}$ is the perturbed prediction relationship introduced by augmentations. $W_{aug}$ changes the original graph label with a flipping probability $p_{aug}$, making $I(W_{inv}\cdot G_{inv}^{e})\neq I(W_{aug}\cdot G_{aug}^{e})$. $N_{e}$ is the independent noise within the training graphs. $\oplus$ is the XOR operation to summarize the impacts of augmentations and noise on the graph label.
With the SEM model in Eqn.~\ref{scm}, we could compute the risk $R$ of any classifier $W$ following prior work \cite{ahuja2021invariance}:
\begin{equation}
\begin{aligned}
R= \mathrm{E}_{e}\mathrm{E}_{Y^{e}_{Aug},G_{Aug}^{e}}[Y^{e}_{Aug}\oplus I(W\cdot G_{Aug}^{e})]
\end{aligned}
\label{risk}
\end{equation}
It is straightforward to verify that the label-invariant augmentation ($p_{aug}=0$)
can guide the GNN to leverage the invariant predictive relationship by risk minimization. 
However, when $p_{aug}\neq0$, the invariant predictive relationship could be sub-optimal, making the GNN classifier to leverage the perturbed predictive relationship.
\begin{theorem}
Denote the risk of $W=W_{inv}$ and $W=W_{aug}$ as $R_{inv}$ and $R_{aug}$ respectively. We have $R_{inv}\geq R_{aug}$ when $p_{aug}\in [\frac{0.5-q}{1-q},1]$ and $R_{inv}< R_{aug}$ when $p_{aug}\in [0,\frac{0.5-q}{1-q})$.
\end{theorem}
The proof is in the appendix. When the label shift occurs in augmentation, a GNN classifier may fail to generalize by leveraging a perturbed predictive relationship. 
In this case, the OOD generalization will be unsatisfactory.
Thus, it is important to maintain label-invariance in graph augmentation for OOD generalization.

\section{Label-Invariant Subgraph Augmentation}

In this work, we seek label-invariant augmentations to enhance the graph OOD generalization performance.

\theoremstyle{definition}
\begin{definition}[Label-invariance Augmentation]
Denote $g\in G$ as the augmentation function and $f:G\rightarrow Y$ as the labeling function. $g$ is a label-invariant augmentation function i.f.f. $f(G)=f(g(G))$.
\end{definition}

\begin{figure*}
\centering
\includegraphics[width=\textwidth]{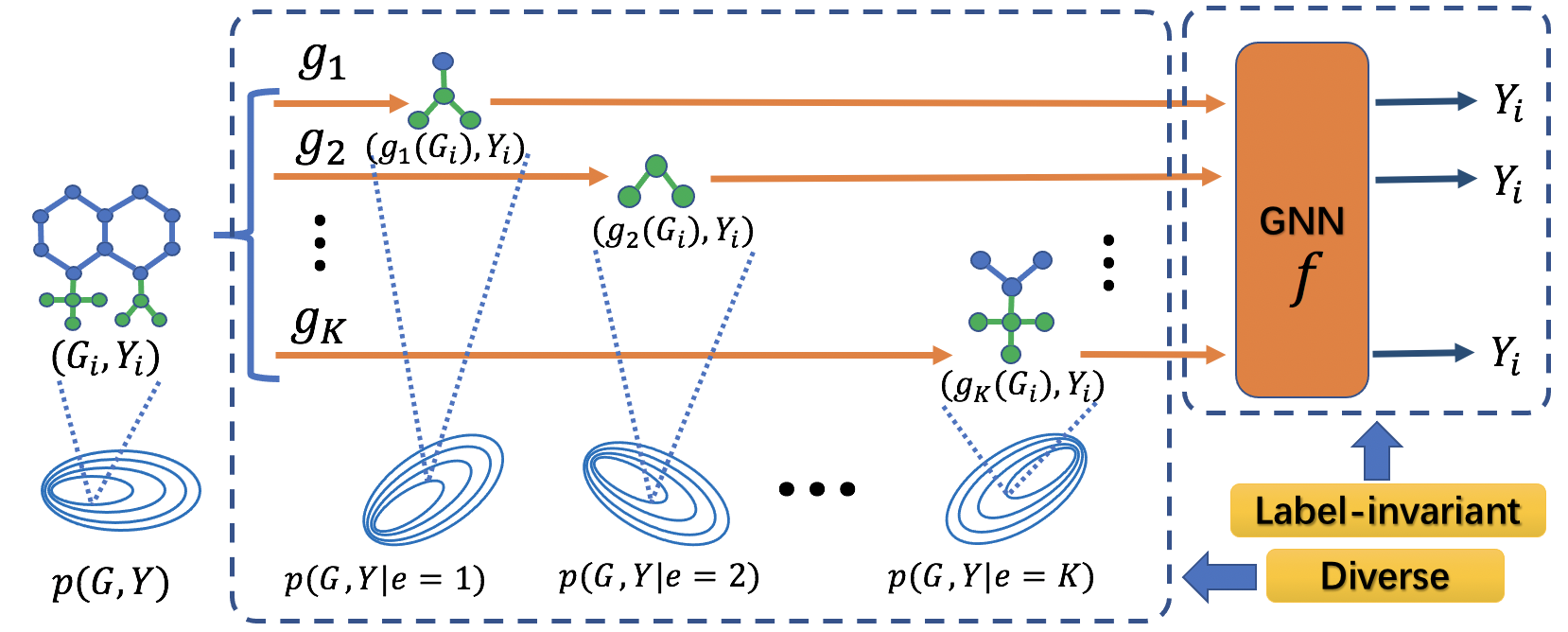}
\vspace{-0.6cm}
\caption{The whole framework of LiSA. LiSA obtains augmented environments by discovering label-invariant subgraphs with a set of variational subgraph generators $\{g_{i}\}_{i=1}^{K}$. Moreover, LiSA employs a tractable energy-based regularization to promote diversity among augmentations. With these environments, LiSA learns an invariant GNN for OOD generalization.}
\label{flowchart}
\vspace{-0.6cm}
\end{figure*}

Designing a label-invariant graph edition policy is usually difficult since it usually needs to model the class-conditioned distribution for graphs $p(G|y)$. 
This usually requires learning a power conditional generative model to simulate $p(G|y)$, which introduces extensive computational burden \cite{jin2020graph-structure}.
For graph-structured data, it is rather difficult to learn such a generative model due to the discrete and vast graph space \cite{jin2018junction}. 
To alleviate this issue, we propose the label-invariant subgraph (LiSA) augmentation method as shown in Figure~\ref{flowchart}. 
Instead of resorting to graph edition, LiSA efficiently generates label-invariant augmentations by exploiting label-invariant subgraphs of the training graphs.

\subsection{Discover Label-invariant Subgraph with Variational Subgraph Generator}

Discovering the label-invariant subgraph of the input graph is non-trivial since the subgraph space is exponentially large \cite{yu2021recognizing}.
We devise the variational subgraph generator to efficiently construct the label-invariant subgraph with a collection of locally predictive patterns. This factorization reduces the subgraph generation into selecting important nodes and edges to avoid directly searching in the large subgraph space.

Given an input graph $G=\{A, X\}$, the variational subgraph generator first outputs a node sampling mask to distill the structure information of $G$.
Specifically, it employs a $l$-layer GNN and a Multi-Layer Perceptron (MLP) to output the sampling probability $p_{v}$ of node $v$:
\begin{equation}
\begin{aligned}
H=\mathrm{GNN}(A,X),\ 
p_{v}=\mathrm{Sigmoid}(\mathrm{MLP}(h_{v})).
\end{aligned}
\label{probability}
\end{equation}
Here, $H$ is the node embedding matrix and $h_{v}$ is the embedding of node $v$. The output of MLP is mapped into [0,1] via the Sigmoid function. 
A large sampling probability guides the node sampling mask $m_{v}=1$ with a high probability, which indicates that the corresponding node $v$ is important to the graph label.
Since the node sampling process is non-differentiable, we further employ the concrete relaxation \cite{jang2016categorical,gal2017concrete} for $m_{v}$:
\begin{equation}
\begin{aligned}
\hat{m}_{v} = \mathrm{Sigmoid}(\frac{1}{t}\log{\frac{p_{v,g_{i}}}{1-p_{v,g_{i}}}}+\log{\frac{u}{1-u}}),
\end{aligned}
\end{equation}
where $t$ is the temperature parameter and $u\sim \mathrm{Uniform}(0,1)$. With the node sampling masks, we further obtain the edge sampling mask by averaging the adjacent nodes. For example, given two adjacent nodes $v$ and $n$, the mask $m_{e}$ for edge $e_{vn}$ is computed as $m_{e}=0.5(m_{v}+m_{n})$. Finally, we mask the input graph $G$ with the node and edge mask to generate the subgraph $G_{sub}$. By introducing the node sampling process, we decompose the subgraph generation process into the node sampling process, which greatly reduces computational expenses. We employ the information constraint \cite{yu2022improving,he2009robust} to restrict that the subgraph only contains a portion of original structural information.
\begin{equation}
\begin{aligned}
\mathcal{L}_{info}&=\mathrm{I}(G,G_{sub})\\
&=\mathbb{E}_{G,G_{sub}} \log{\frac{p(G_{sub}|G)}{q(G_{sub})}}-\mathrm{KL}[p(G_{sub})|q(G_{sub})]\\
&\leq \mathbb{E}_{G\sim p(G)} \mathrm{KL}[p(G_{sub}|G)|G)|q(G_{sub})].
\end{aligned}
\label{inequality}
\end{equation}
Here, $\mathrm{KL}$ is the KL-divergence. The inequality is due to the fact that KL-divergence is non-negative. The  posterior distribution $p(G_{sub}|G,e=i)$ is factorized into $\prod_{v\in G} \mathrm{Bernoulli}(p_{v})$ due to the node sampling process.
The specification of the prior $p(G_{sub}|e=i)$ in Eqn.~\ref{inequality} is the non-informative distribution $\prod_{v\in G}\mathrm{Bernoulli}(0.5)$ following \cite{alemi2016deep}, which encodes equal probability of sampling or dropping nodes in prior knowledge. 

We proceed to encode the label-invariance into the obtained subgraph with a jointly trained GNN classifier $f$. In each iteration, it is first updated with the labeled training graphs. Then, it serves as a proxy to measure the gap between the graph label and the subgraph label.
\begin{equation}
\begin{aligned}
\mathcal{L}_{cls} &= \mathrm{CE}(f(g(G)),Y),
\end{aligned}
\label{predictable}
\end{equation}
where $\mathrm{CE}$ is the cross-entropy loss. During the subgraph generation process, the variational subgraph generator recognizes the label-invariant subgraph by jointly minimizing the following loss:
\begin{equation}
\begin{aligned}
\mathcal{L} &= \mathcal{L}_{cls}(f,g) + \alpha\mathcal{L}_{info}(g).
\end{aligned}
\label{loss-generator}
\end{equation}

\subsection{Graph OOD Generalization with LiSA}
Existing works show that the performance drop in graph OOD generalization results from preferring superficial knowledge, such as spurious subgraphs, for the prediction \cite{wu2022discovering,chen2022invariance}. 
With locally easy-to-learn information, the GNN classifier can achieve a low training risk without a global understanding of the whole graph structure, making it difficult for OOD generalization. 
Our work alleviates this issue by first decomposing each training graph into multiple label-invariant subgraphs using a set of variational subgraph generators.
Intuitively, different subgraph generators generate diverse label-invariant subgraphs to construct the augmented training environments. Suppose we use $n$ variational subgraph generators; we can obtain $n+1$ training environments together with the original training graphs. We treat the subgraphs produced by the same variational subgraph generator as an augmented environment. And the probability density function of $n+1$ total environments is $p(G,Y)=\frac{1}{n+1}\sum_{i=1}^{n+1}p(G,Y|e_{i})$ and $e$ is the environment variable. 
Then, we aim to train an invariant GNN which performs equally well on these environments. 
In this manner, the GNN avoids from only preferring the spurious subgraph for the prediction and give stable prediction on different locally crucial patterns.

Denote the GNN classifier as $f$. We employ the variance regularization \cite{krueger2021out} to learn an invariant GNN:
\begin{equation}
\begin{aligned}
\min_{f} \mathcal{L}_{cls}(f) + \mathrm{Var}_{e}(\mathcal{L}_{cls}(f)).
\end{aligned}
\label{loss-cls}
\end{equation}
The first term is the classification loss of GNN on all the training environments and the second term is the variance of classification losses in different environments. To jointly optimize the variational subgraph generators and the GNN classifier, we minimize the loss terms in Eqn.~\ref{loss-generator} and Eqn.~\ref{loss-cls} with a bi-level optimization framework:
\begin{equation}
\begin{aligned}
&\min_{f} \mathcal{L}_{cls}(f,g_{i}^{*}) + \mathrm{Var}_{e}(\mathcal{L}_{cls}(f,g_{i}^{*})),i=1\sim n\\
&s.t.g_{i}^{*}=\arg\min_{g_{i}}\mathcal{L}_{cls}(f,g_{i})+\alpha \mathcal{L}_{info}(g_{i}).
\end{aligned}
\label{loss-bilevel}
\end{equation}
In practice, we first obtain a sub-optimal $g^{*}$ by optimizing $g$ for $T$ steps in the inner loop. Then, we use the updated $g^{*}$ as a proxy in the outer loop to optimize $f$. We provide pseudo-code for optimizing Eqn.~\ref{loss-bilevel} in Appendix.

\subsection{Enforcing Diversity Among Augmented environments}
Directly optimizing Eqn.~\ref{loss-bilevel} may lead to a sub-optimal solution where
different variational subgraph generators generate similar subgraphs. Thus, we aim to enlarge the distances between the distributions of different augmented environments to promote diversity.

\begin{table*}[htbp]
  \centering
  \setlength{\tabcolsep}{1.1mm}{
  \caption{Performances of different methods on OOD graph classification tasks. We report the mean and standard deviation of Accuracy. In the Spurious-Motif dataset, $b$ is the indicator of spurious correlation.}
  \vspace{-0.2cm}
    \begin{tabular}{c|c|c|c|c|c|c|c}
    \toprule
    \multirow{2}[4]{*}{Methods} & \multicolumn{4}{c|}{SpuriousMotif} & \multirow{2}[4]{*}{MUTAG} & \multirow{2}[4]{*}{MNIST-75sp} & \multirow{2}[4]{*}{DD} \\
\cmidrule{2-5}          & 0.33  & 0.5   & 0.7   & 0.9   &       &       &  \\
    \midrule
    ERM   & 0.509 $\pm$ 0.007 & 0.505 $\pm$ 0.004 & 0.490 $\pm$ 0.006 & 0.448 $\pm$ 0.004 & 0.903 $\pm$ 0.009 & 0.862 $\pm$ 0.015 & 0.718 $\pm$ 0.027 \\
    IRM   & 0.502 $\pm$ 0.003 & 0.501 $\pm$ 0.005 & 0.486 $\pm$ 0.007 & 0.443 $\pm$ 0.017 & 0.910 $\pm$ 0.015 & 0.875 $\pm$ 0.006 & 0.732 $\pm$ 0.017 \\
    V-Rex & 0.526 $\pm$ 0.010 & 0.518 $\pm$ 0.010 & 0.484 $\pm$ 0.010 & 0.452 $\pm$ 0.017 & 0.900 $\pm$ 0.020 & 0.868 $\pm$ 0.006 & 0.730 $\pm$ 0.031 \\
    Attention & 0.514 $\pm$ 0.038 & 0.484 $\pm$ 0.045 & 0.452 $\pm$ 0.049 & 0.430 $\pm$ 0.016 & 0.917 $\pm$ 0.012 & 0.878 $\pm$ 0.003 & 0.529 $\pm$ 0.053 \\
    TopKPool & 0.439 $\pm$ 0.028 & 0.432 $\pm$ 0.038 & 0.482 $\pm$ 0.035 & 0.366 $\pm$ 0.006 & 0.913 $\pm$ 0.007 & \textbf{0.879 $\pm$ 0.003} & 0.663 $\pm$ 0.031 \\
    GIB   & 0.524 $\pm$ 0.024 & 0.492 $\pm$ 0.019 & 0.430 $\pm$ 0.062 & 0.355 $\pm$ 0.003 & 0.887 $\pm$ 0.053 & 0.865 $\pm$ 0.002 & 0.543 $\pm$ 0.178 \\
    DIR   & 0.468 $\pm$ 0.025 & 0.459 $\pm$ 0.030 & 0.427 $\pm$ 0.021 & 0.386 $\pm$ 0.011 & 0.895 $\pm$ 0.049 & 0.812 $\pm$ 0.031 & 0.741 $\pm$ 0.074 \\
    \midrule
    \textbf{LiSA} & \textbf{0.530 $\pm$ 0.004} & \textbf{0.529 $\pm$ 0.003} & \textbf{0.501 $\pm$ 0.005} & \textbf{0.474 $\pm$ 0.009} & \textbf{0.937 $\pm$ 0.014} & 0.876 $\pm$ 0.008 & \textbf{0.746 $\pm$ 0.069} \\
    \bottomrule
    \end{tabular}
    \label{graph-level}}
    \vspace{-0.5cm}
\end{table*}%

\textbf{Energy-based Regularization.}
We propose a novel energy-based diversity regularization to enlarge the distance between the underlying distributions of augmented environments.
We employ the energy-based model (EBM) $p(G|e) \propto \exp{-E_{\theta}(G|e)}$ to specify the graph distribution. Here $E_{\theta}:\mathcal{G}\rightarrow \mathcal{R}$ is the energy score and $\theta$ is the model parameter. 
The energy score assigns the density of data points in each environment. Thus, we can compute the distance between the distributions of two environments based on the energy scores of pair-wised samples:
\begin{equation}
\begin{aligned}
d(e_{j},e_{k})= \frac{1}{2N}\sum_{i=1}^{N}[E_{\theta_{j}}(G_{i}|e_{j}),E_{\theta_{k}}(G_{i}|e_{k})]^{2}.
\end{aligned}
\label{energy-score}
\end{equation}
Directly computing the distance in Eqn.~\ref{energy-score} requires estimating the model parameters $\theta_{j}$ and $\theta_{k}$ of EBMs in two environments, which is computationally inefficient. To this end, we compute the energy scores with the predictive logits of the GNN classifier $f$. Recall that the GNN classifier outputs the prediction by applying the Softmax function to the predictive logits.
\begin{equation}
\begin{aligned}
p(Y|G,e_{j})&=\frac{\exp{f(g_{j}(G))[Y]}}{\sum_{Y\in\mathcal{Y}}\exp{f(g_{j}(G))[Y]}},
\end{aligned}
\label{eq8}
\end{equation}
where $f()[Y]$ denotes the $Y$-th output of $f$. Following prior work \cite{grathwohl2019your}, we obtain the joint distribution $p(Y,G|e_{j})= \frac{\exp{f(g_{j}(G))[Y]}}{Z}$. Here $Z$ is the partition function. Then, we marginalize $Y$ to obtain $p(G|e_{j})=\frac{\sum_{Y\in\mathcal{Y}}\exp{f(g_{j}(G))[Y]}}{Z}$. Combining Eqn.~\ref{eq8}, the energy score is expressed using the predictive logits:
\begin{equation}
\begin{aligned}
E_{\theta_{j}}(G|e_{j})=-\log\sum_{Y\in\mathcal{Y}}\exp{f(g_{j}(G))[Y]}.
\end{aligned}
\label{energy}
\end{equation}
Combining Eqn.~\ref{energy} and Eqn.~\ref{energy-score}, we can compute pair-wise distances among environments with the energy score:
\begin{equation}
\begin{aligned}
\mathcal{L}_{e}=\frac{2}{N(N+1)}\sum_{j=1}^{N}\sum_{k=j+1}^{N+1}d(e_{j},e_{k}).
\end{aligned}
\label{energy-dist}
\end{equation}
Thus, the total loss of LiSA takes the following form:
\begin{equation}
\begin{aligned}
&\min_{f} \mathcal{L}_{cls}(f,\{g_{i}^{*}\}_{i=1}^{n}) + \mathrm{Var}_{e}(\mathcal{L}_{cls}(f,g_{i}^{*})),i=1\sim n\\
&s.t.g_{i}^{*}=\arg\min_{g_{i}}\mathcal{L}_{cls}(f,g_{i})+\alpha \mathcal{L}_{info}(g_{i})+\beta \mathcal{L}_{e}(g_{i}).
\end{aligned}
\label{lisa-v}
\end{equation}

\subsection{Extension to Node-level Tasks}
We proceed to introduce the extension of LiSA on node classification tasks. Different from the graph classification task,  the nodes are associated with their neighborhoods in the node classification task. Hence, we take a local view of the nodes and relate them with 1-hop ego-graphs \cite{wu2022handling,zhu2021transfer}. For example, $N_{i}$ is associated with $G_{i}=(A_{i}, X_{i})$, where $A_{i}$ is the adjacent matrix of the 1-hop subgraph centered at $N_{i}$ and $X_{i}$ is the neighborhood node feature matrix. 
Then, we generate multiple label-invariant subgraphs of $G_{i}$ by optimizing the subgraph generators with Eqn.~\ref{loss-generator}. 
The whole framework of LiSA is optimized with Eqn.~\ref{lisa-v}.

\section{Experiments}
\label{section-experiment}
In this section, we extensively evaluate LiSA on both node-level and graph-level OOD generalization tasks with different types of distribution shifts. We run experiments on the server with Tesla V100 GPU and Intel(R) Xeon(R) Gold 6348 CPU, and use the PyG for implementation. The network architecture, sensitivity study of hyper-parameters, and detailed information on datasets are in the appendix.

\subsection{Graph-level OOD Generalization}
We first evaluate LiSA on out-of-distribution (OOD) graph classification tasks with various distribution shifts such as the graph size, noise feature, and spurious motif.

\textbf{Datasets.} We employ \textbf{Spurious-Motif} \cite{gnnexplainer}, \textbf{MUTAG} \cite{nr}, \textbf{D\&D} \cite{conf/nips/KnyazevTA19}, and \textbf{MNIST-75sp} \cite{conf/nips/KnyazevTA19} datasets for OOD graph classification. The Spurious-Motif dataset consists of synthetic graphs with spurious motifs. Each graph is generated by attaching one base (Tree, Ladder, Wheel, denoted as $S=0,1,2$) to a motif (Cycle, House, Crane, denoted as $C=0,1,2$). The graph label $Y$ is consistent with the class of motif. For the training graphs, the base is chosen with probability $P(S)=b\times\mathds{1}(S=C) + \frac{1-b}{2}\times\mathds{1}(S\neq C)$ to create a spurious correlation. $b$ is changed to impose different biases on the training graphs. For testing graphs, the motifs and bases are randomly connected. The training and testing data in D\&D and MUTAG datasets vary in the graph size. Specifically, we choose the graphs in the D\&D dataset with less than 200 nodes for training, those with 200-300 nodes for validation, and graphs larger than 300 nodes for testing. For MUTAG, we select graphs with less than 15 nodes for training, those with 15-20 nodes for validation, and graphs larger than 20 nodes for testing. For MNIST-75sp, each image is converted as super-pixel graphs. The features of testing data contain random noises. We report accuracy (Acc) for these datasets.

\begin{figure*}
\centering
\vspace{-0.5cm}
\includegraphics[width=\textwidth]{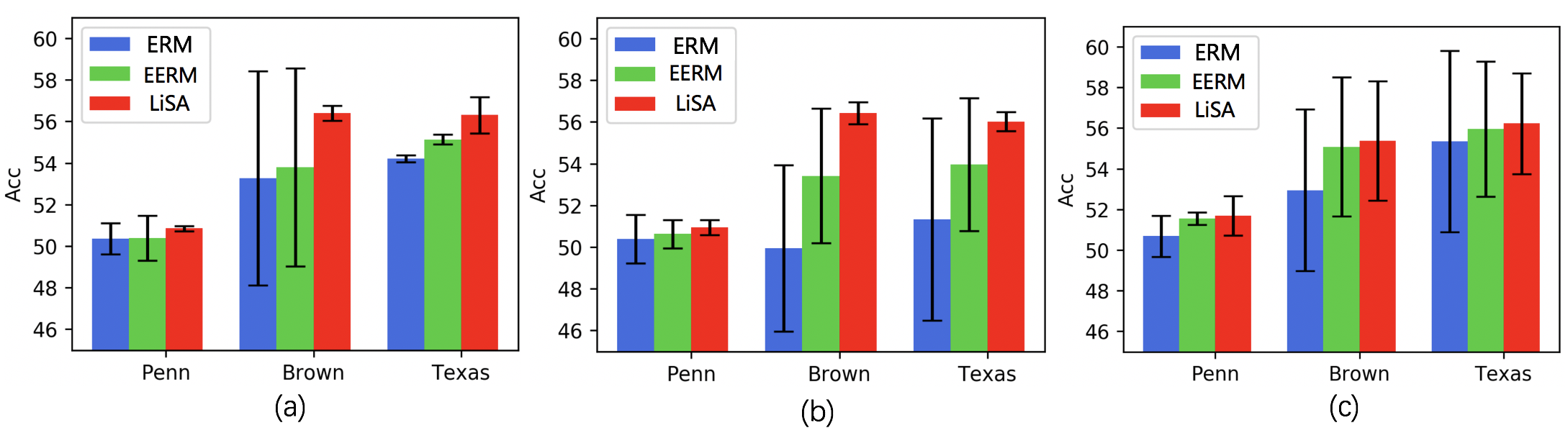}
\vspace{-0.7cm}
\centering
\caption{We report performances of different methods with 3 source domain combinations on the Facebook-100 dataset: (a). John Hopkins + Caltech + Amherst; (b).Bingham + Duke + Princeton; and (c). WashU + Brandeis+ Carnegie. We report the mean and standard deviation of Accuracy across different runs.}
\vspace{-0.5cm}
\label{facebook}
\end{figure*}

\begin{table*}[htbp]
  \centering
  \caption{Mean and standard deviation of Accuracy (Acc) on OGB-Arxiv dataset.}
    \begin{tabular}{c|c|c|c|c|c|c}
    \toprule
    Test Domain & \multicolumn{2}{c|}{14-16} & \multicolumn{2}{c|}{16-18} & \multicolumn{2}{c}{18-20} \\
    \midrule
    Backbone & APPNP & SGGCN & APPNP & SGGCN & \multicolumn{1}{c|}{APPNP} & \multicolumn{1}{c}{SGGCN} \\
    \midrule
    ERM   & \multicolumn{1}{l|}{46.30 $\pm$ 0.35} & \multicolumn{1}{l|}{40.52 $\pm$ 1.24} & \multicolumn{1}{l|}{43.75 $\pm$ 0.40} & \multicolumn{1}{l|}{38.23 $\pm$ 2.15} & 39.78 $\pm$ 0.41 & 34.62 $\pm$ 2.14 \\
    EERM  & \multicolumn{1}{l|}{46.42 $\pm$ 0.46} & \multicolumn{1}{l|}{42.37 $\pm$  2.37} & \multicolumn{1}{l|}{44.53 $\pm$ 0.54} & \multicolumn{1}{l|}{39.91 $\pm$ 2.07} & \textbf{43.24 $\pm$ 0.79} & 37.73 $\pm$ 1.42 \\
    \textbf{LiSA} & \textbf{47.50 $\pm$ 0.52} & \textbf{47.14 $\pm$ 0.34} & \textbf{45.10 $\pm$ 0.50} & \textbf{45.49 $\pm$ 0.37} & 41.216 $\pm$ 0.249 & \textbf{38.89 $\pm$ 0.71} \\
    \bottomrule
    \end{tabular}%
  \label{arxiv}%
  \vspace{-0.6cm}
\end{table*}%

\begin{table*}[htbp]
  \centering
  \caption{Test ROC-AUC on Twitch-Explicit dataset. Each method is trained in a single environment. For a fair comparison, both EERM and LiSA generate 3 augmented environments.}
  \vspace{-0.4cm}
    \begin{tabular}{c|c|c|c|c|c}
    \toprule
    GCN   & ES    & FR    & PTBR  & RU    & TW \\
    \midrule
    ERM   & 52.50 $\pm$ 4.09 & 54.92 $\pm$ 2.60 & 48.78 $\pm$ 7.45 & 50.49 $\pm$ 1.82 & 48.95 $\pm$ 2.31 \\
    EERM  & 54.17 $\pm$ 5.04 & 54.10 $\pm$ 1.76 & 49.49 $\pm$ 7.96 & 51.34 $\pm$ 1.672 & 49.83 $\pm$ 3.15 \\
    LiSA-Rex & 57.75 $\pm$ 3.75 & 53.77 $\pm$ 0.84 & 55.40 $\pm$ 9.04 & 52.47 $\pm$ 0.39 & \textbf{54.66 $\pm$ 0.53} \\
    \textbf{LiSA} & \textbf{57.97 $\pm$ 2.96} & \textbf{55.87 $\pm$ 2.66} & \textbf{59.96 $\pm$ 2.12} & \textbf{52.73 $\pm$ 0.67} & 52.60 $\pm$ 2.64 
    \\
    LiSA w/o $\mathcal{L}_{e}$ & 57.28 $\pm$ 3.49 & 54.80 $\pm$ 1.37 & 57.73 $\pm$ 7.23 & 52.55 $\pm$ 0.93 & 52.67 $\pm$ 2.21 \\
    LiSA w/o $\mathcal{L}_{info}$ & 55.81 $\pm$ 2.21 & 54.94 $\pm$ 2.49 & 57.49 $\pm$ 2.17 & 51.76 $\pm$ 0.91 & 50.71 $\pm$ 2.47 \\
    \bottomrule
    \end{tabular}%
  \label{twitter}%
  \vspace{-0.5cm}
\end{table*}%

\textbf{Baselines.} We compare our method with empirical risk minimization (ERM), invariant learning methods, including V-Rex \cite{krueger2021out} and IRM \cite{arjovsky2019invariant}; interpretable methods, such as Attention-based Pooling \cite{conf/nips/KnyazevTA19}, TopK-Pooling \cite{gao2019graph}, and GIB \cite{yu2020graph}; and augmentation-based invariant learning method DIR \cite{wu2022discovering}.
We employ GIN \cite{Xu:2019ty} as the backbone for model-agnostic baselines.
Since there is only one domain for training, we randomly group graphs to mimic different domains to instantiate V-Rex and IRM. We report the mean and standard deviation of testing performances in 10 runs for different methods.

\textbf{Performance.} 
We report graph OOD classification performance in Table~\ref{graph-level}. LiSA outperforms most baseline methods on both synthetic and real-world datasets, with up to 5\% absolute performance gain. When compared with ERM, IRM and V-Rex only achieve comparable performance on most datasets. This shows that it is usually insufficient to directly implement general OOD generalization methods to handle the distribution shifts on graphs. For interpretable methods, they somehow achieve performance gain compared with ERM as they discover a prediction-relevant subgraph for predictions. However, they underperform other methods when a large distribution shift occurs in the dataset, such as $b=0.9$ for Spurious Motif and DD. Moreover, we find that DIR, an augmentation-based invariant learning method for graphs, can sometimes underperform ERM or interpretable methods due to the label shift problem in augmentation. Thus, it is important to maintain label invariance during augmentation.

\subsection{Node-level OOD Generalization}
We proceed to apply LiSA to the OOD node classification, where the distribution shifts are spatial and temporal. 

\textbf{Datasets \& Metrics.} For the spatial shift, we adopt \textbf{Twitch-Explicit} \cite{rozemberczki2021multi} and \textbf{Facebook-100} \cite{traud2012social} datasets for evaluation. These datasets contain different social networks which are related to different locations such as campuses and districts. For example, Twitch-Explicit contains seven social networks, including DE, ENGB, ES, FR, PTBR, RU, and TW. Following the protocol in prior work \cite{wu2022handling}, we employ DE for training, ENGB for validation, and the rest five networks for testing. For the Facebook-100 dataset, we choose different combinations of three graphs for training, two for validation, and the rest three graphs for testing. We report ROC-AUC and Accuracy (Acc) for Twitch-Explicit and Facebook-100.

For the temporal shift, we use a citation network \textbf{OGB-Arxiv} \cite{hu2020open}, and a dynamic financial dataset \textbf{ELLIPTIC} \cite{pareja2020evolvegcn}. For OGB-Arxiv, we employ the papers published before 2011 for training, from 2011$\sim$2014 for validation, and those within 2014$\sim$2016/2016$\sim$2018/2018$\sim$2020 for testing. For ELLIPTIC, we split the whole dataset into different snapshots, and use 5/5/33 for training, validation, and testing. The testing environments are further chronologically clustered into 9 folders for the convenience of comparing the performances of different methods. We report Test F1 Score and Accuracy (Acc) for ELLIPTIC and OGB-Arxiv.

\textbf{Baselines.} We compare the performance of the proposed LiSA with \textbf{ERM} and the state-of-the-art node generalization method, Explore-to-Extrapolate Risk Minimization (\textbf{EERM}) \cite{wu2022handling}. EERM generates augmentations by adding new edges while LiSA generates diverse label-invariant subgraphs. For a fair comparison, we generate 3 augmented domains for both EERM and LiSA. 
We further plug different methods into various GNN backbones, such as GCN \cite{gcn}, GraphSAGE \cite{conf/nips/HamiltonYL17}, APPNP \cite{klicpera2018predict}, SGGCN \cite{wu2019simplifying} and GCNII \cite{chen2020simple}, to extensively evaluate their performance. 
We evaluate the model with the highest validation accuracy and report the mean and standard deviation of 10-run performance for each method.

\textbf{Performance.} We report the results on Twitch-Explicit in Table~\ref{twitter}. The proposed LiSA exceeds the baselines in most testing environments. Since there is only one source domain for training, ERM is difficult to generalize. EERM outperforms ERM in 4 of 5 testing environments. 
In Figure~\ref{facebook}, we compare different methods with the GCN backbone on the Facebook-100 dataset. We can see that LiSA achieves performance gains in different training environments.
Moreover, the performance variance of LiSA is low, showing a more stable generalization performance compared with EERM and ERM. 

\begin{figure}[t]
\begin{center}
\centerline{\includegraphics[width=0.5\textwidth]{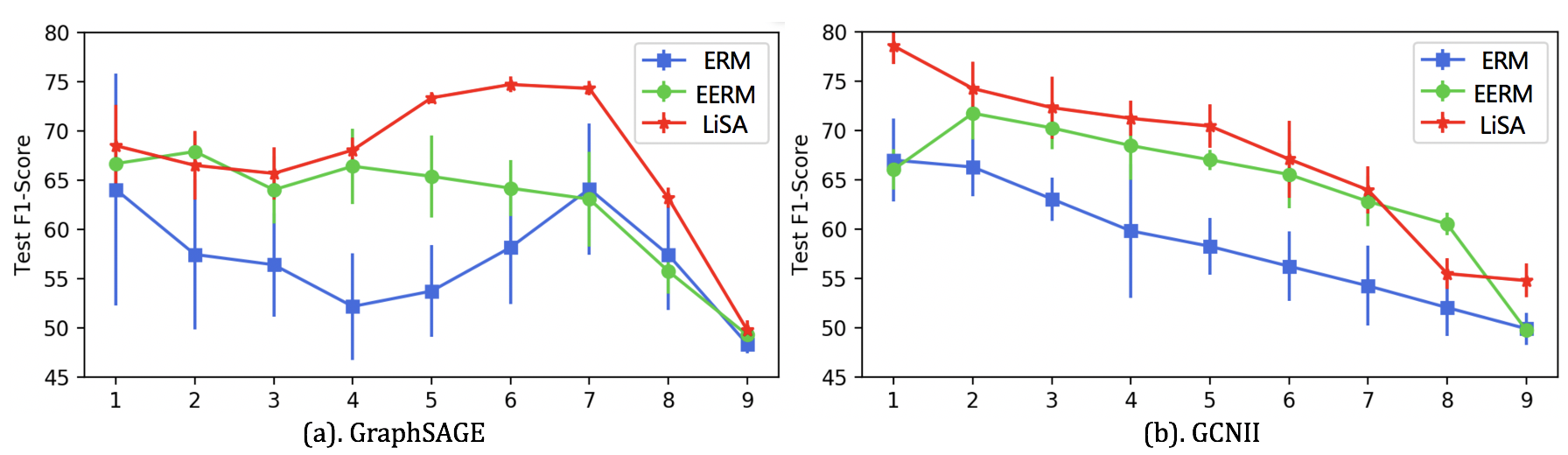}}
\end{center}
\vspace{-0.8cm}
\caption{Test F1-Score on 9 testing folds in the ELLIPTIC dataset. LiSA achieves better OOD generalization performance with different GNN backbones.}
\label{elliptic}
\vspace{-0.7cm}
\end{figure}

For the temporal shift on nodes, we first plug different methods into APPNP and SGGCN backbones and evaluate their performances on the OGB-Arxiv dataset. As shown in Table~\ref{arxiv}, LiSA outperforms the baselines in five cases out of six with stable results in different runs. 
Then, we report the results on the Elliptic dataset in Figure~\ref{elliptic}. LiSA outperforms the baseline methods in most testing folds with different backbones and achieves up to 10\% absolute performance gain. 
Moreover, we observe using different backbones can lead to different generalization performances. Nevertheless, LiSA still achieves better generalization performances when using the same backbone as the baselines.

\subsection{Discussions}
\textbf{Influence of Backbone on OOD Generalization.} The OOD generalization performance on graphs is sensitive to GNN backbones. As shown in Table~\ref{arxiv}, using APPNP as the backbone usually results in better generalization performance for the node-level temporal shift. Moreover, the generalization performances in Figure~\ref{elliptic} behave differently when choosing different backbones. Thus, apart from label-invariant augmentation, adopting an appropriate GNN backbone is also essential for graph OOD generalization.

\textbf{Ablation Study.} We remove $\mathcal{L}_{info}$ and $\mathcal{L}_{e}$ from Eqn.~\ref{lisa-v} to study their effects on label-invariant augmentation. As shown in Table~\ref{twitter}, removing these terms leads to performance drops in generalization, which validates that these two terms are important to generate label-invariant subgraphs for augmentation. Notice that prior work \cite{wu2022handling} maximizes the variance of classifier in the augmented environments to promote diversity. We replace $\mathcal{L}_{e}$ in Eqn.~\ref{lisa-v} with the variance-based regularization, leading to LiSA-Rex. As shown in Table~\ref{twitter}, LiSA-Rex also achieves competitive performance. LiSA-Rex outperforms EERM when they both employ variance-based regularization for diversity. Thus, it is important to maintain label-invariant augmentation for graph OOD generalization. Moreover, using the proposed energy-based regularization leads to better OOD generalization performance than variance-based regularization since LiSA outperforms LiSA-Rex.

\textbf{On the Diversity of Augmentations.} Moreover, we study the augmentation performances of different methods. We compare the distance between the original training environment and 3 augmented environments, denoted as $d_{1}\sim d_{3}$, and the average pair-wise distance across 3 augmented environments, denoted as $d_{intra}$. We employ the score-based distance in OOD detection \cite{liu2020energy} as the distance metric. As shown in Table~\ref{distance}, the augmented environments generated by EERM are similar since they have a small average pair-wise distance, which shows insufficient diversity. 
LiSA produces augmented environments with a large average pair-wise distance. Moreover, the augmented domains are also far from the source domain. Hence, LiSA can indeed generate more diverse augmentations to facilitate graph OOD generalization.

\begin{table}
  \centering
  \caption{On the diversity of different augmentation methods. $d_{1}\sim d_{3}$ are distances between the original training environment and the augmented environments. $d_{intra}$ is the average pair-wise distance across augmented environments.}
  \vspace{-0.3cm}
    \begin{tabular}{c|cccc}
    \toprule
      Distance    & $d_{1}$   & $d_{2}$   & $d_{3}$   & $d_{intra}$ \\
    \midrule
    EERM  & 0.76  & 0.73  & 0.75  & 0.04 \\
    LiSA   & 0.67  & 0.70   & 0.64  & 0.52 \\
    \bottomrule
    \end{tabular}%
  \label{distance}%
  \vspace{-0.6cm}
\end{table}%

\section{Conclusion}
In this work, we have studied augmentation-based OOD generalization for graphs. We show that the label shift during augmentation makes the learned GNN hard to generalize, and thus it is crucial to maintain label-invariant augmentation. We propose LiSA to efficiently generate  label-invariant augmentations by exploiting multiple label-invariant subgraphs of the training graphs. LiSA contains a set of variational subgraph generators to discover label-invariant subgraphs efficiently. Moreover, a novel energy-based regularization is proposed to promote diversity among augmentations. With these augmentations, LiSA can learn an invariant GNN that is expected to generalize.
LiSA is model-agnostic and can be plugged into various GNN backbones.
Extensive experiments on node-level and graph-level benchmarks show the superior performance of LiSA on various graph OOD generalization tasks. 

\section*{Acknowledgement}
This work is partially funded by the National Natural Science Foundation of China (Grant No. U21B2045), the National Natural Science Foundation of China (Grant No. U20A20223), the Youth Innovation Promotion Association CAS (Grant No. Y201929), and Beijing Nova Program under (Grant Z211100002121108).
{\small
\bibliographystyle{ieee_fullname}
\bibliography{egbib}

\begin{thebibliography}{10}\itemsep=-1pt

\bibitem{ahuja2021invariance}
Kartik Ahuja, Ethan Caballero, Dinghuai Zhang, Jean-Christophe Gagnon-Audet,
  Yoshua Bengio, Ioannis Mitliagkas, and Irina Rish.
\newblock Invariance principle meets information bottleneck for
  out-of-distribution generalization.
\newblock {\em Advances in Neural Information Processing Systems},
  34:3438--3450, 2021.

\bibitem{alemi2016deep}
Alexander~A Alemi, Ian Fischer, Joshua~V Dillon, and Kevin Murphy.
\newblock Deep variational information bottleneck.
\newblock {\em arXiv preprint arXiv:1612.00410}, 2016.

\bibitem{arjovsky2019invariant}
Martin Arjovsky, L{\'e}on Bottou, Ishaan Gulrajani, and David Lopez-Paz.
\newblock Invariant risk minimization.
\newblock {\em arXiv preprint arXiv:1907.02893}, 2019.

\bibitem{baranwal2021graph}
Aseem Baranwal, Kimon Fountoulakis, and Aukosh Jagannath.
\newblock Graph convolution for semi-supervised classification: Improved linear
  separability and out-of-distribution generalization.
\newblock {\em arXiv preprint arXiv:2102.06966}, 2021.

\bibitem{bevilacqua2021size}
Beatrice Bevilacqua, Yangze Zhou, and Bruno Ribeiro.
\newblock Size-invariant graph representations for graph classification
  extrapolations.
\newblock In {\em International Conference on Machine Learning}, pages
  837--851. PMLR, 2021.

\bibitem{bian2020rumor}
Tian Bian, Xi Xiao, Tingyang Xu, Peilin Zhao, Wenbing Huang, Yu Rong, and
  Junzhou Huang.
\newblock Rumor detection on social media with bi-directional graph
  convolutional networks.
\newblock In {\em Proceedings of the AAAI conference on artificial
  intelligence}, pages 549--556, 2020.

\bibitem{buffelli2022sizeshiftreg}
Davide Buffelli, Pietro Li{\`o}, and Fabio Vandin.
\newblock Sizeshiftreg: a regularization method for improving
  size-generalization in graph neural networks.
\newblock {\em arXiv preprint arXiv:2207.07888}, 2022.

\bibitem{buhlmann2020invariance}
Peter B{\"u}hlmann.
\newblock Invariance, causality and robustness.
\newblock {\em Statistical Science}, 35(3):404--426, 2020.

\bibitem{chang2021not}
Heng Chang, Yu Rong, Tingyang Xu, Yatao Bian, Shiji Zhou, Xin Wang, Junzhou
  Huang, and Wenwu Zhu.
\newblock Not all low-pass filters are robust in graph convolutional networks.
\newblock {\em Advances in Neural Information Processing Systems},
  34:25058--25071, 2021.

\bibitem{chang2022adversarial}
Heng Chang, Yu Rong, Tingyang Xu, Wenbing Huang, Honglei Zhang, Peng Cui, Xin
  Wang, Wenwu Zhu, and Junzhou Huang.
\newblock Adversarial attack framework on graph embedding models with limited
  knowledge.
\newblock {\em IEEE Transactions on Knowledge and Data Engineering (TKDE)},
  2022.

\bibitem{chang2020restricted}
Heng Chang, Yu Rong, Tingyang Xu, Wenbing Huang, Honglei Zhang, Peng Cui, Wenwu
  Zhu, and Junzhou Huang.
\newblock A restricted black-box adversarial framework towards attacking graph
  embedding models.
\newblock In {\em Proceedings of the AAAI Conference on Artificial
  Intelligence}, volume~34, pages 3389--3396, 2020.

\bibitem{chen2020simple}
Ming Chen, Zhewei Wei, Zengfeng Huang, Bolin Ding, and Yaliang Li.
\newblock Simple and deep graph convolutional networks.
\newblock In {\em International Conference on Machine Learning}, pages
  1725--1735. PMLR, 2020.

\bibitem{chen2022invariance}
Yongqiang Chen, Yonggang Zhang, Han Yang, Kaili Ma, Binghui Xie, Tongliang Liu,
  Bo Han, and James Cheng.
\newblock Invariance principle meets out-of-distribution generalization on
  graphs.
\newblock {\em arXiv preprint arXiv:2202.05441}, 2022.

\bibitem{chen2023pareto}
Yongqiang Chen, Kaiwen Zhou, Yatao Bian, Binghui Xie, Bingzhe Wu, Yonggang
  Zhang, MA KAILI, Han Yang, Peilin Zhao, Bo Han, and James Cheng.
\newblock Pareto invariant risk minimization: Towards mitigating the
  optimization dilemma in out-of-distribution generalization.
\newblock In {\em The Eleventh International Conference on Learning
  Representations}, 2023.

\bibitem{chuang2022tree}
Ching-Yao Chuang and Stefanie Jegelka.
\newblock Tree mover's distance: Bridging graph metrics and stability of graph
  neural networks.
\newblock {\em arXiv preprint arXiv:2210.01906}, 2022.

\bibitem{creager2021environment}
Elliot Creager, J{\"o}rn-Henrik Jacobsen, and Richard Zemel.
\newblock Environment inference for invariant learning.
\newblock In {\em International Conference on Machine Learning}, pages
  2189--2200. PMLR, 2021.

\bibitem{cubuk2018autoaugment}
Ekin~D Cubuk, Barret Zoph, Dandelion Mane, Vijay Vasudevan, and Quoc~V Le.
\newblock Autoaugment: Learning augmentation policies from data.
\newblock {\em arXiv preprint arXiv:1805.09501}, 2018.

\bibitem{fischer2021stickypillars}
Kai Fischer, Martin Simon, Florian Olsner, Stefan Milz, Horst-Michael Gross,
  and Patrick Mader.
\newblock Stickypillars: Robust and efficient feature matching on point clouds
  using graph neural networks.
\newblock In {\em Proceedings of the IEEE/CVF Conference on Computer Vision and
  Pattern Recognition}, pages 313--323, 2021.

\bibitem{gal2017concrete}
Yarin Gal, Jiri Hron, and Alex Kendall.
\newblock Concrete dropout.
\newblock {\em arXiv preprint arXiv:1705.07832}, 2017.

\bibitem{gao2019graph}
Hongyang Gao and Shuiwang Ji.
\newblock Graph u-nets.
\newblock In {\em international conference on machine learning}, pages
  2083--2092. PMLR, 2019.

\bibitem{grathwohl2019your}
Will Grathwohl, Kuan-Chieh Wang, J{\"o}rn-Henrik Jacobsen, David Duvenaud,
  Mohammad Norouzi, and Kevin Swersky.
\newblock Your classifier is secretly an energy based model and you should
  treat it like one.
\newblock {\em arXiv preprint arXiv:1912.03263}, 2019.

\bibitem{conf/nips/HamiltonYL17}
William~L. Hamilton, Zhitao Ying, and Jure Leskovec.
\newblock Inductive representation learning on large graphs.
\newblock In {\em Advances in neural information processing systems}, pages
  1024--1034, 2017.

\bibitem{he2009robust}
Ran He, Bao-Gang Hu, and Xiao-Tong Yuan.
\newblock Robust discriminant analysis based on nonparametric maximum entropy.
\newblock In {\em Advances in Machine Learning: First Asian Conference on
  Machine Learning, ACML 2009, Nanjing, China, November 2-4, 2009. Proceedings
  1}, pages 120--134. Springer, 2009.

\bibitem{heinze2018invariant}
Christina Heinze-Deml, Jonas Peters, and Nicolai Meinshausen.
\newblock Invariant causal prediction for nonlinear models.
\newblock {\em Journal of Causal Inference}, 6(2), 2018.

\bibitem{hu2020open}
Weihua Hu, Matthias Fey, Marinka Zitnik, Yuxiao Dong, Hongyu Ren, Bowen Liu,
  Michele Catasta, and Jure Leskovec.
\newblock Open graph benchmark: Datasets for machine learning on graphs.
\newblock {\em Advances in neural information processing systems},
  33:22118--22133, 2020.

\bibitem{hu2018does}
Weihua Hu, Gang Niu, Issei Sato, and Masashi Sugiyama.
\newblock Does distributionally robust supervised learning give robust
  classifiers?
\newblock In {\em International Conference on Machine Learning}, pages
  2029--2037. PMLR, 2018.

\bibitem{jang2016categorical}
Eric Jang, Shixiang Gu, and Ben Poole.
\newblock Categorical reparameterization with gumbel-softmax.
\newblock {\em arXiv preprint arXiv:1611.01144}, 2016.

\bibitem{jiang2022invariant}
Yibo Jiang and Victor Veitch.
\newblock Invariant and transportable representations for anti-causal domain
  shifts.
\newblock {\em arXiv preprint arXiv:2207.01603}, 2022.

\bibitem{jin2018junction}
Wengong Jin, Regina Barzilay, and Tommi Jaakkola.
\newblock Junction tree variational autoencoder for molecular graph generation.
\newblock In {\em International Conference on Machine Learning}, pages
  2323--2332. PMLR, 2018.

\bibitem{jin2020multi}
Wengong Jin, Regina Barzilay, and Tommi Jaakkola.
\newblock Multi-objective molecule generation using interpretable
  substructures.
\newblock In {\em International Conference on Machine Learning}, pages
  4849--4859. PMLR, 2020.

\bibitem{jin2020graph-structure}
Wei Jin, Yao Ma, Xiaorui Liu, Xianfeng Tang, Suhang Wang, and Jiliang Tang.
\newblock Graph structure learning for robust graph neural networks.
\newblock In {\em Proceedings of the 26th ACM SIGKDD International Conference
  on Knowledge Discovery \& Data Mining}, pages 66--74, 2020.

\bibitem{khanam2022homophily}
Kazi~Zainab Khanam, Gautam Srivastava, and Vijay Mago.
\newblock The homophily principle in social network analysis: A survey.
\newblock {\em Multimedia Tools and Applications}, pages 1--44, 2022.

\bibitem{gcn}
Thomas~N. Kipf and Max Welling.
\newblock Semi-supervised classification with graph convolutional networks.
\newblock In {\em The International Conference on Representation Learning},
  2017.

\bibitem{kleinberg2018algorithmic}
Jon Kleinberg, Jens Ludwig, Sendhil Mullainathan, and Ashesh Rambachan.
\newblock Algorithmic fairness.
\newblock In {\em Aea papers and proceedings}, volume 108, pages 22--27, 2018.

\bibitem{klicpera2018predict}
Johannes Klicpera, Aleksandar Bojchevski, and Stephan G{\"u}nnemann.
\newblock Predict then propagate: Graph neural networks meet personalized
  pagerank.
\newblock {\em arXiv preprint arXiv:1810.05997}, 2018.

\bibitem{conf/nips/KnyazevTA19}
Boris Knyazev, Graham~W. Taylor, and Mohamed~R. Amer.
\newblock Understanding attention and generalization in graph neural networks.
\newblock In {\em NeurIPS}, pages 4204--4214, 2019.

\bibitem{koyama2020out}
Masanori Koyama and Shoichiro Yamaguchi.
\newblock Out-of-distribution generalization with maximal invariant predictor.
\newblock 2020.

\bibitem{krueger2021out}
David Krueger, Ethan Caballero, Joern-Henrik Jacobsen, Amy Zhang, Jonathan
  Binas, Dinghuai Zhang, Remi Le~Priol, and Aaron Courville.
\newblock Out-of-distribution generalization via risk extrapolation (rex).
\newblock In {\em International Conference on Machine Learning}, pages
  5815--5826. PMLR, 2021.

\bibitem{li2022out}
Haoyang Li, Xin Wang, Ziwei Zhang, and Wenwu Zhu.
\newblock Out-of-distribution generalization on graphs: A survey.
\newblock {\em arXiv preprint arXiv:2202.07987}, 2022.

\bibitem{liu2020energy}
Weitang Liu, Xiaoyun Wang, John Owens, and Yixuan Li.
\newblock Energy-based out-of-distribution detection.
\newblock {\em Advances in Neural Information Processing Systems},
  33:21464--21475, 2020.

\bibitem{pareja2020evolvegcn}
Aldo Pareja, Giacomo Domeniconi, Jie Chen, Tengfei Ma, Toyotaro Suzumura,
  Hiroki Kanezashi, Tim Kaler, Tao Schardl, and Charles Leiserson.
\newblock Evolvegcn: Evolving graph convolutional networks for dynamic graphs.
\newblock In {\em Proceedings of the AAAI Conference on Artificial
  Intelligence}, volume~34, pages 5363--5370, 2020.

\bibitem{peters2016causal}
Jonas Peters, Peter B{\"u}hlmann, and Nicolai Meinshausen.
\newblock Causal inference by using invariant prediction: identification and
  confidence intervals.
\newblock {\em Journal of the Royal Statistical Society: Series B (Statistical
  Methodology)}, 78(5):947--1012, 2016.

\bibitem{qian2019robust}
Qi Qian, Shenghuo Zhu, Jiasheng Tang, Rong Jin, Baigui Sun, and Hao Li.
\newblock Robust optimization over multiple domains.
\newblock In {\em Proceedings of the AAAI Conference on Artificial
  Intelligence}, pages 4739--4746, 2019.

\bibitem{nr}
Ryan~A. Rossi and Nesreen~K. Ahmed.
\newblock The network data repository with interactive graph analytics and
  visualization.
\newblock In {\em AAAI}, 2015.

\bibitem{rozemberczki2021multi}
Benedek Rozemberczki, Carl Allen, and Rik Sarkar.
\newblock Multi-scale attributed node embedding.
\newblock {\em Journal of Complex Networks}, 9(2):cnab014, 2021.

\bibitem{sagawa2019distributionally}
Shiori Sagawa, Pang~Wei Koh, Tatsunori~B Hashimoto, and Percy Liang.
\newblock Distributionally robust neural networks for group shifts: On the
  importance of regularization for worst-case generalization.
\newblock {\em arXiv preprint arXiv:1911.08731}, 2019.

\bibitem{shen2021towards}
Zheyan Shen, Jiashuo Liu, Yue He, Xingxuan Zhang, Renzhe Xu, Han Yu, and Peng
  Cui.
\newblock Towards out-of-distribution generalization: A survey.
\newblock {\em arXiv preprint arXiv:2108.13624}, 2021.

\bibitem{shorten2019survey}
Connor Shorten and Taghi~M Khoshgoftaar.
\newblock A survey on image data augmentation for deep learning.
\newblock {\em Journal of big data}, 6(1):1--48, 2019.

\bibitem{sinha2017certifying}
Aman Sinha, Hongseok Namkoong, Riccardo Volpi, and John Duchi.
\newblock Certifying some distributional robustness with principled adversarial
  training.
\newblock {\em arXiv preprint arXiv:1710.10571}, 2017.

\bibitem{tailor2021towards}
Shyam~A Tailor, Ren{\'e} de Jong, Tiago Azevedo, Matthew Mattina, and Partha
  Maji.
\newblock Towards efficient point cloud graph neural networks through
  architectural simplification.
\newblock In {\em Proceedings of the IEEE/CVF International Conference on
  Computer Vision}, pages 2095--2104, 2021.

\bibitem{traud2012social}
Amanda~L Traud, Peter~J Mucha, and Mason~A Porter.
\newblock Social structure of facebook networks.
\newblock {\em Physica A: Statistical Mechanics and its Applications},
  391(16):4165--4180, 2012.

\bibitem{volpi2018generalizing}
Riccardo Volpi, Hongseok Namkoong, Ozan Sener, John~C Duchi, Vittorio Murino,
  and Silvio Savarese.
\newblock Generalizing to unseen domains via adversarial data augmentation.
\newblock {\em Advances in neural information processing systems}, 31, 2018.

\bibitem{wu2022survey}
Bingzhe Wu, Jintang Li, Junchi Yu, Yatao Bian, Hengtong Zhang, CHaochao Chen,
  Chengbin Hou, Guoji Fu, Liang Chen, Tingyang Xu, et~al.
\newblock A survey of trustworthy graph learning: Reliability, explainability,
  and privacy protection.
\newblock {\em arXiv preprint arXiv:2205.10014}, 2022.

\bibitem{wu2019simplifying}
Felix Wu, Amauri Souza, Tianyi Zhang, Christopher Fifty, Tao Yu, and Kilian
  Weinberger.
\newblock Simplifying graph convolutional networks.
\newblock In {\em International conference on machine learning}, pages
  6861--6871. PMLR, 2019.

\bibitem{wu2022handling}
Qitian Wu, Hengrui Zhang, Junchi Yan, and David Wipf.
\newblock Handling distribution shifts on graphs: An invariance perspective.
\newblock {\em arXiv preprint arXiv:2202.02466}, 2022.

\bibitem{wu2020graph}
Shiwen Wu, Fei Sun, Wentao Zhang, and Bin Cui.
\newblock Graph neural networks in recommender systems: a survey.
\newblock {\em arXiv preprint arXiv:2011.02260}, 2020.

\bibitem{wu2022discovering}
Ying-Xin Wu, Xiang Wang, An Zhang, Xiangnan He, and Tat-Seng Chua.
\newblock Discovering invariant rationales for graph neural networks.
\newblock {\em International Conference on Learning Representations}, 2022.

\bibitem{Xu:2019ty}
Keyulu Xu, Weihua Hu, Jure Leskovec, and Stefanie Jegelka.
\newblock {How Powerful are Graph Neural Networks?}
\newblock In {\em Proceedings of the 7th International Conference on Learning
  Representations}, ICLR '19, pages 1--17, 2019.

\bibitem{gnnexplainer}
Rex Ying, Dylan Bourgeois, Jiaxuan You, Marinka Zitnik, and Jure Leskovec.
\newblock Gnnexplainer: Generating explanations for graph neural networks.
\newblock In {\em Advances in neural information processing systems}, 2019.

\bibitem{yu2022improving}
Junchi Yu, Jie Cao, and Ran He.
\newblock Improving subgraph recognition with variational graph information
  bottleneck.
\newblock {\em IEEE Conferences on Computer Vision and Pattern Recognition},
  2022.

\bibitem{yu2020graph}
Junchi Yu, Tingyang Xu, Yu Rong, Yatao Bian, Junzhou Huang, and Ran He.
\newblock Graph information bottleneck for subgraph recognition.
\newblock {\em International Conference on Learning Representations}, 2021.

\bibitem{yu2021recognizing}
Junchi Yu, Tingyang Xu, Yu Rong, Yatao Bian, Junzhou Huang, and Ran He.
\newblock Recognizing predictive substructures with subgraph information
  bottleneck.
\newblock {\em IEEE Transations on Pattern Analysis and Machine Intelligence},
  2021.

\bibitem{yu2022structure}
Junchi Yu, Tingyang Xu, Yu Rong, Junzhou Huang, and Ran He.
\newblock Structure-aware conditional variational auto-encoder for constrained
  molecule optimization.
\newblock {\em Pattern Recognition}, 126:108581, 2022.

\bibitem{zhang2022correct}
Michael Zhang, Nimit~S Sohoni, Hongyang~R Zhang, Chelsea Finn, and Christopher
  R{\'e}.
\newblock Correct-n-contrast: A contrastive approach for improving robustness
  to spurious correlations.
\newblock {\em arXiv preprint arXiv:2203.01517}, 2022.

\bibitem{zhou2022leverage}
Mengxi Zhou, Wei Xu, Wenping Zhang, and Qiqi Jiang.
\newblock Leverage knowledge graph and gcn for fine-grained-level clickbait
  detection.
\newblock {\em World Wide Web}, 25(3):1243--1258, 2022.

\bibitem{zhu2021transfer}
Qi Zhu, Carl Yang, Yidan Xu, Haonan Wang, Chao Zhang, and Jiawei Han.
\newblock Transfer learning of graph neural networks with ego-graph information
  maximization.
\newblock {\em Advances in Neural Information Processing Systems}, 34, 2021.

\end{thebibliography}
}

\end{document}